*Data and text mining*

# Pre-training technique to localize medical BERT and enhance biomedical BERT

Shoya Wada[1,*], Toshihiro Takeda[1], Shiro Manabe[1], Shozo Konishi[1],

Jun Kamohara[2], and Yasushi Matsumura[1]

[1] Department of Medical Informatics, Osaka University Graduate School of Medicine, 2-2 Yamadaoka, Suita, Osaka, Japan, [2] Faculty of Medicine, Osaka University, 2-2 Yamadaoka, Suita, Osaka, Japan.

*To whom correspondence should be addressed.

**Abstract**
**Background:** Pre-training large-scale neural language models on raw texts has made a significant contribution to improving transfer learning in natural language processing (NLP). With the introduction of transformer-based language models, such as bidirectional encoder representations from transformers (BERT), the performance of information extraction from a free text by NLP has significantly improved for both the general domain and medical domain; however, it is difficult to train specific BERT models that perform well for domains in which there are few publicly available databases of high quality and large size. We hypothesized that this problem can be addressed by up-sampling a domain-specific corpus and using it for pre-training with a larger corpus in a balanced manner.
**Method:** Our proposed method consists of a single intervention with one option: simultaneous pre-training after up-sampling, which is intended to encourage masked language modeling and next-sentence prediction on the small corpus, and amplified vocabulary, which helps to suit the smaller corpus when building the customized vocabulary by byte-pair encoding. We conducted three experiments and evaluated the resulting products; (1) Japanese medical BERT from a small medical corpus, (2) English biomedical BERT from a small biomedical corpus, and (3) Enhanced biomedical BERT from whole PubMed abstracts.
**Results:** We confirmed that our Japanese medical BERT outperformed conventional baselines and the other BERT models in terms of the medical document classification task and that our English BERT pre-trained using both the general and medical-domain corpora performed sufficiently well for practical use in terms of the biomedical language understanding evaluation (BLUE) benchmark. Moreover, our enhanced biomedical BERT model, in which clinical notes were not used during pre-training, showed that both the clinical and biomedical scores of the BLUE benchmark were 0.3 points above that of the ablation model trained without our proposed method.
**Conclusions:** Well-balanced pre-training by up-sampling instances derived from a corpus appropriate for the target task allows us to construct a high-performance BERT model.

## 1   Background

Pre-training large-scale neural language models on raw texts has proven to considerably improve the process of transfer learning in natural language processing (NLP). With the introduction of transformer-based language models, such as bidirectional encoder representations from transformers (BERT), the information extraction performance of NLP from a free text has significantly improved in the general domain [11,12]. Meanwhile, due to the rapid increase in the volume of medical literature, it is expected that the accuracy of information extraction in the biomedical domain will improve as well. Many studies, such as those on BioBERT, clinicalBERT, and SciBERT, have shown that pre-training BERT models on a large domain-specific text corpus, such as biomedical, clinical, or scientific text, results in satisfactory performance in their specific text-mining tasks [13-15].

However, with regard to localization of medical BERT models, significant barriers continue to exist today. There are few publicly available medical databases written in languages other than English with high quality and a large size sufficient to train BERT models. For example, in Japanese, a subscription is required to perform a cross-search of Japanese medical journals, and most articles are published only in the PDF format, making it difficult to obtain a large medical corpus. Thus, there is a high demand for techniques to build language models that work well even when the resources available are limited. In this regard, certain data augmentation techniques have been proposed for NLP [18]; however, there are no reports on how up-sampling affects the pre-training of BERT.

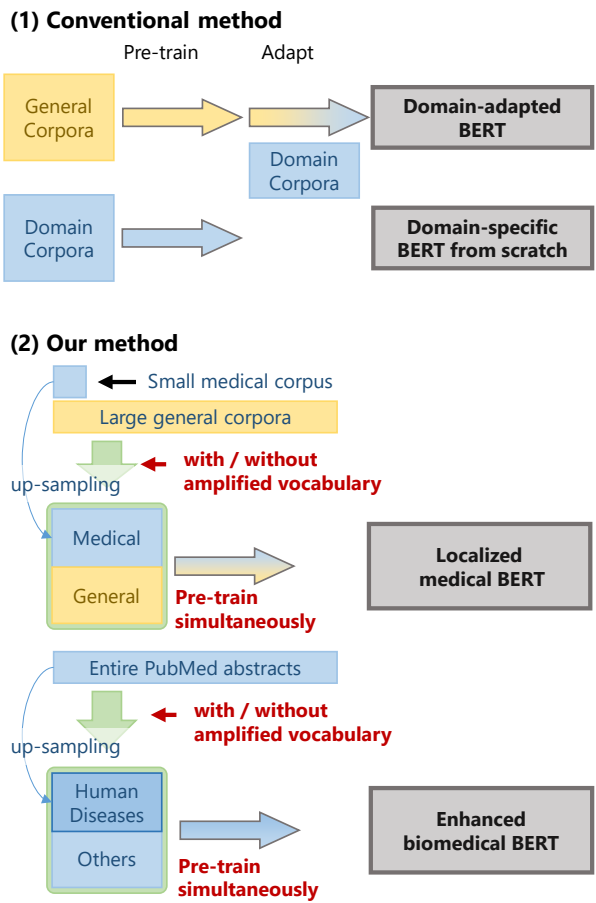

Figure 1. Overview of pre-training BERT.

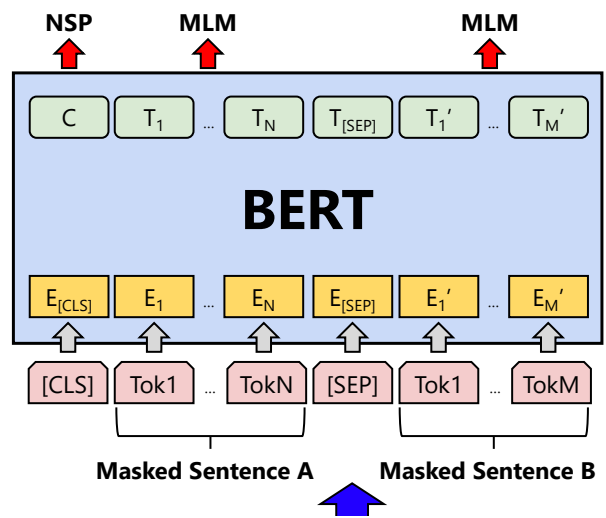

Figure 2. Pre-training procedures for BERT (adapted from Devlin 2019).
The input instances consist of two sentences, such as text spans separated by a special token [SEP]. [CLS] is a special instance marker added in front of each input example and used for next-sentence prediction (NSP). The tokens replaced with [MASK] are used for masked language modeling (MLM).

We hypothesized that the above-mentioned problem could be solved by up-sampling a domain-specific corpus and using it for pre-training in a balanced manner with a larger corpus. In this paper, we describe our method and demonstrate that it can process an objective task with higher performance.

We propose simultaneous pre-training after up-sampling in which we distinguish between two types of corpora and combine them to create pre-training instances via our method. Accordingly, we have also developed appropriate BERT models (see Figure 1). In this paper, we first show the improvement that our method offers over conventional models for a medical document classification task. Second, we confirm whether our method can be applied to other domains. We apply our method to an English-based model and verify that the performance of the model is comparable to that of models built using a conventional method. Third, we demonstrate that our approach enables the development of a pre-trained model that enhances biomedical BERT in both clinical and biomedical tasks by balancing the corpora used for pre-training.

In particular, our study makes the following contributions:

(1) We propose a method that enables users to train a medical BERT model using a small corpus. Subsequently, we show that the localization of medical BERT is feasible using our method.

(2) Applying our method, we developed a pre-trained BERT model using only PubMed abstracts and released it for biomedical text mining by Osaka University (ouBioBERT). We compared the performance of ouBioBERT with the existing BERT models and achieve higher performance in clinical tasks in terms of the biomedical language understanding evaluation (BLUE) benchmark [14].

## 2 Methods

Our models essentially have the same structures as those of BERT-Base. Therefore, we begin this section with an overview of BERT and describe the available models used for medical text-mining tasks. Next, we describe our method and refer to our models. Finally, we explain the fine-tuning process to evaluate our models.

To validate our method, we conducted three experiments (see Section 2.3):

(1) Japanese medical BERT from a small medical corpus
(2) English biomedical BERT from a small biomedical corpus
(3) Enhanced biomedical BERT from whole PubMed abstracts

### 2.1 BERT: Bidirectional encoder representations from transformers

BERT [12] is a contextualized word-representation model based on masked language modeling (MLM) that is pre-trained using bidirectional transformers [11]. The BERT framework consists of two steps: pre-training and fine-tuning. During pre-training, the model is trained on unlabeled large corpora. For fine-tuning, the BERT model is first initialized with pre-trained weights, and all the weights are fine-tuned using labeled data from

**Table 1:** Comparison of common medical terms in vocabularies used by BERT variants.

| Medical Term | Category | BERT | SciBERT | ouBioBERT (Ours) |
|---|---|---|---|---|
| stroke | disease | ✓ | ✓ | ✓ |
| malalia | disease | ✓ | ✓ | ✓ |
| bleeding | symptom | ✓ | ✓ | ✓ |
| seizure | symptom | ✓ | ✓ | ✓ |
| pulmonary | organ | ✓ | ✓ | ✓ |
| stomach | organ | ✓ | ✓ | ✓ |
| surgery | procedure | ✓ | ✓ | ✓ |
| prescription | procedure | ✓ | ✓ | ✓ |
| cocaine | chemical | ✓ | ✓ | ✓ |
| glucose | chemical | ✓ | ✓ | ✓ |
| osteoporosis | disease |  | ✓ | ✓ |
| edema | symptom |  | ✓ | ✓ |
| pancreas | organ |  | ✓ | ✓ |
| laparotomy | procedure |  | ✓ | ✓ |
| dexamethasone | chemical |  | ✓ | ✓ |
| appendicitis | disease |  |  | ✓ |
| jaundice | symptom |  |  | ✓ |
| duodenum | organ |  |  | ✓ |
| polypectomy | procedure |  |  | ✓ |
| codeine | chemical |  |  | ✓ |

Note: a "✓" symbol indicates that the corresponding vocabulary contains the medical term; otherwise, the term will be broken up into smaller subwords.

the downstream tasks. We applied minimal architectural modifications to the task-specific inputs and outputs into BERT and fine-tuned all the parameters in an end-to-end manner.

#### 2.1.1 Pre-training

BERT pre-training is optimized for two unsupervised classification tasks (Figure 2). The first is MLM. One training instance of MLM is a single modified sentence. Each token in the sentence has a 15% chance of being replaced by a special token [MASK]. The chosen token is replaced with [MASK] 80% of the time, 10% with another random token, and the remaining 10% with the same token. The MLM objective is a cross-entropy loss on predicting the masked tokens.

The second task is next-sentence prediction (NSP), which is a binary classification loss for predicting whether two segments follow each other in the original text. Positive instances are created by taking consecutive sentences from the text corpus. Negative instances are created by pairing segments from different documents. Positive and negative instances are sampled with equal probabilities. The NSP objective is designed to improve the performance of downstream tasks, such as natural language inference (NLI) [22], which requires reasoning regarding the relationships between pairs of sentences.

While creating the training instances, we set a duplicate factor, which contributes to data augmentation, while pre-training BERT. It refers to the duplicating times of the instances created from an input sentence, where these instances originate from the same sentence but have different [MASK] tokens (see lines 38-43 in Figure 3).

#### 2.1.2 Vocabulary

To manage the problem of out-of-vocabulary words, BERT uses vocabulary from subword units generated by WordPiece [23], which is based on byte-pair encoding (BPE) [24], for the unsupervised tokenization of the input text. The vocabulary is built such that it contains the most frequently used words or subword units. The main benefit of pre-training from scratch is to leverage a domain-specific custom vocabulary. Table 1 compares the vocabularies used by BERT variants. For example, *appendicitis*, a common disease name, is divided into four pieces ([app, ##end, ##ici, ##tis]) by BERT [12] and three pieces ([append, ##icit, ##is]) by SciBERT [25].

#### 2.1.3 Pre-trained BERT variants

The standard BERT model has been reported to not perform well in specialized domains, such as biomedical or scientific texts [13,25]. To overcome this limitation, there are two possible strategies: either additional pre-training on domain-specific corpora from an existing pre-trained BERT model or pre-training from scratch on domain-specific corpora. The main benefit of the former is that the computational cost of pre-training is lower than that of the latter. The main advantage of the latter, as mentioned, is the availability of its custom vocabulary; however, the disadvantage is that the pre-trained neural language model may be less adaptable if the number of documents in a specific domain is small.

**BERT-Base** was pre-trained using English Wikipedia (2,500M words) and BooksCorpus (800M words) [12]. The vocabulary is BaseVocab, and its size is 30 K. Some published models have been initialized from BERT-Base and trained using their domain-specific corpora.

**BlueBERT** was published with the BLUE benchmark [14]. In this study, we evaluated BlueBERT-Base (P) and BlueBERT-Base (P + M), which were initialized from BERT-Base and additionally pre-trained using only PubMed abstracts and using a combination of PubMed abstracts for 5M steps and MIMIC-III clinical notes [26] for 0.2M steps, respectively. We refer to them as biomedical BlueBERT and clinical BlueBERT, respectively.

**Tohoku-BERT** is a Japanese BERT model used for the general domain released by Tohoku University [27]. It was pre-trained using Japanese Wikipedia, and its vocabulary was obtained by applying BPE to the corpus.

**UTH-BERT** is a clinical BERT model in Japan published by the University of Tokyo [28]. It was developed using a vast volume of Japanese clinical narrative text, and its vocabulary was built with consideration of segment words for diseases or findings in as large a unit as possible.

```
1   Scorpus = SmallCorpus
2   Lcorpus = LargeCorpus
3
4   tokenizer = BertWordPieceTokenizer()
5   model = BertForPreTraining()
6
7   # Create tokenizer
8   if args.AmpV == True:
9     tokenizer.train(
10      corpus=(Scorpus * int(Lcorpus.filesize
11                          / Scorpus.filesize)
12            + Lcorpus)
13    )
14  else:
15    tokenizer.train(
16      corpus=(Scorpus + Lcorpus)
17    )
18
19  # Create pre-training instances
20  if args.SimPT == True:
21    split_Sdocs = split_corpus(corpus=Scorpus,
22                               each_file_size=10MB)
23    split_Ldocs = split_corpus(corpus=Lcorpus,
24                               each_file_size=10MB)
25    for _ in range(args.n_docs):
26      # Sampling evenly from the split documents
27      docs = (random.sample(split_Sdocs, 10)
28             + random.sample(split_Ldocs, 10))
29      instances.extend(
30        create_instances_from_document(
31          documents=docs,
32          tokenizer=tokenizer)
33      )
34  else:
35    split_Alldocs = split_corpus(corpus=(Scorpus+Lcorpus),
36                                 n_docs=args.n_docs)
37    for docs in split_Alldocs:
38      for _ in range(args.dupe_factor):
39        instances.extend(
40          create_instances_from_document(
41            documents=docs,
42            tokenizer=tokenizer)
43        )
44
45  # Train
46  model.train(instances)
```

**Figure 3.** Pseudocode for our proposed method.
AmpV: amplified vocabulary; SimPT: simultaneous pre-training after up-sampling; dupe_factor: duplicate factor. *Scorpus* and *Lcorpus* indicate corpora which contain multiple documents. *tokenizer* is used to tokenize an input sentence. The *split_corpus* function splits the target corpus into the specified file size or number of splits. The *create_instances_from_document* function creates training instances from documents.

1. Select whether to use AmpV or not and create a tokenizer to process texts (lines 7-17).
2. Create pre-training instances. If SimPT is used, run lines 21-32 (our proposed method); otherwise, run lines 34-42 (the original implementation).
3. Use the instances to train a model (line 46).

### 2.2 Proposed method: simultaneous pre-training after up-sampling and amplified vocabulary

It is generally known that if we train a BERT model only on a small medical corpus, there is a possibility that overfitting may degrade its performance. We hypothesized that this issue can be avoided if we simultaneously train a BERT model using knowledge from both general as well as the medical domains. This can be achieved by increasing the frequency of pre-training for MLM using documents of the medical domain rather than the general domain and using the negative instances of NSP in which a sentence pair is constructed by pairing two random sentences, each from a different document. To increase the number of combinations of documents and to enhance medical-word representations in the vocabulary, we introduce the following intervention with one option:

**Simultaneous pre-training after up-sampling** (hereinafter, referred to as "SimPT") is a technique used to efficiently create pre-training instances from a set of corpora according to their file sizes and to pre-train a neural language model, as described in Figure 3. In the case of a medical BERT model, a small corpus corresponds to a medical corpus, and a large corpus is a general domain corpus, such as Wikipedia.

In the original implementation, we first divide the entire corpus into smaller text files that contain multiple documents. Subsequently, the combinations of NSP are determined within each split file, and the duplicate factor is set to define the number of times the sentences are used; however, there are two problems. First, the duplicate factor is applied to the entire corpora of both the small corpus and the large corpus, and thus the smaller corpus remains relatively small in pre-training instances. This means that up-sampling is not possible in this implementation. The second problem is that the combinations of NSP are limited to the file that was initially split (see lines 34-43 in Figure 3).

In our method, both the small and large corpora are first divided into smaller (and different) documents of the same size and then combined to create pre-training instances. When we combine them, we ensure that the documents in the small and large corpora are comparable in terms of their file sizes and that the patterns of the combination are diverse. Using this technique, more instances from the small corpus are used than those in the original implementation (see lines 20-33 in Figure 3). Consequently, this intervention achieves up-sampling of the small corpus. Furthermore, it generates an increased number of different combinations of documents compared to the original method.

As described in Figure 3, documents derived from the small corpus and those derived from the large corpus are combined such that their proportion is equal in terms of their file sizes, and a sufficient number of pre-training instances are created to train the BERT model.

The **amplified vocabulary** (hereinafter referred to as "AmpV") is a custom vocabulary used to suit the small corpus. If we build a vocabulary with BPE without adjusting the corpus file sizes of the small and large corpora, most words and subwords would be derived from the large corpus. To solve this problem, we amplify the small corpus and make the corpus file size the same as that of the large corpus. Subsequently, we construct the uncased vocabulary via BPE using tokenizers [29] (see lines 8-17 in Figure 3). AmpV is an option of SimPT and is used when creating pre-training instances (see lines 29-32 in Figure 3).

### 2.3 Our pre-trained models and experimental settings

We produced the following BERT-Base models to demonstrate the proposed method. The corpora used in our models are listed in Table 2. The difference between our pre-trained models and the published models is clarified in Table A1.

#### 2.3.1 Japanese medical BERT from a small medical corpus

**Our BERT (jpCR + jpW)** is a Japanese medical BERT model that was pre-trained using our method. We used the medical reference, "Today's diagnosis and treatment: premium," which consisted of 15 digital resources for clinicians in Japanese published by IGAKU-SHOIN Ltd. as the source for the medical domain (abbreviated as "jpCR"). Similarly, Japanese Wikipedia (jpW) was used for the general domain.

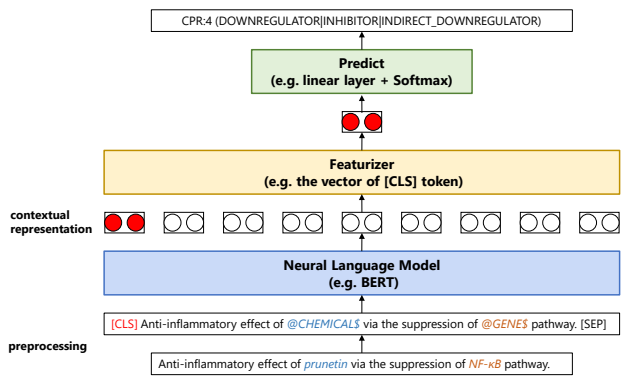

**Figure 4.** A general architecture of task-specific fine-tuning on neural language models; an example of relation-extraction, ChemProt [3].

A sentence is transformed into an instance for BERT by replacing target entities with dummy tokens and adding special tokens. In a relation-extraction task, we use [CLS] BERT encoding as a Featurizer and predict the relationship between the entities by multiclass classification.

Four pre-trained models were prepared for comparison. Two of them are publicly available models: **Tohoku-BERT** and **UTH-BERT**. The others are pre-trained models using conventional methods in our environment: **BERT (jpW/jpCR)**, which was initialized with Tohoku-BERT and trained for additional steps using Japanese clinical references, and **BERT (jpCR)**, which was pre-trained using only Japanese clinical references from scratch.

### 2.3.2 English biomedical BERT from a small biomedical corpus

Evaluation with multiple tasks is desirable; however, in this study, our Japanese model could be evaluated only with a single task. To assess whether there are any disadvantages of up-sampling in BERT's pre-training, we evaluated our method in English as well.

First, we empirically produced a limited corpus of clinically relevant articles from PubMed abstracts. PubMed comprises a large number of citations for biomedical literature from MEDLINE and, therefore, its articles constitute a mix from the fields of clinical medicine and life sciences. We constructed a small biomedical corpus, denoted as "sP," extracted from PubMed abstracts by using their medical subject headings (MeSH) IDs, which can be converted to the corresponding tree number. The heuristic rules used to decide which articles to extract are listed in Table A2. Next, we created a general corpus randomly sampled from articles in English Wikipedia to replicate the experiment in Japanese. It was denoted as "sW," which is similar to Japanese Wikipedia in terms of word count and file size.

**Our BERT (sP + sW)** is the name of the pre-trained medical BERT model in English. We used small PubMed abstracts (sP) as the small medical source and sampled English Wikipedia (sW) as the general corpus. **BERT (sP)** and **BERT (sW/sP)** were trained for comparison. The former was pre-trained solely using sP from scratch, and the latter was initialized from **BERT (sW)**, which was pre-trained using sampled English Wikipedia in our environment, and trained using small PubMed abstracts for domain-specific adaptation similar to BioBERT [13].

**Table 2.** List of the text corpora used for our models.

| Corpus | | Number of words | Size (GB) | Domain |
|---|---|---|---|---|
| (jpW) | Japanese Wikipedia | 550M | 2.6 | (jp) General |
| (jpCR) | Clinical References | 18M | 0.1 | (jp) Medical |
| (sW) | Sampled English Wikipedia | 500M | 3.0 | (en) General |
| (sP) | Small PubMed abstracts | 18M | 0.1 | (en) Biomedical |
| (fP) | Focused PubMed abstracts | 280M | 1.8 | (en) Biomedical |
| (oP) | Other PubMed abstracts | 2,800M | 18 | (en) Biomedical |
| | Entire PubMed abstracts | 3,100M | 20 | (en) Biomedical |

Notes: Japanese corpora are tokenized using MeCab [4]. jp: Japanese; en: English.

### 2.3.3 Enhanced biomedical BERT from whole PubMed abstracts

Previous research has shown that domain-adaptive pre-training is effective and that additional task-adaptive pre-training enhances the performance of downstream tasks [30]. PubMed articles, which are commonly used in biomedical language models, constitute a mix from the fields of clinical medicine and life sciences; however, biomedical NLP tasks are mainly focused on human beings. Therefore, we hypothesized that our approach could boost the amount of training on articles related to human diseases within the entire corpus of PubMed and thus evaluated the effect.

We created focused PubMed abstracts, denoted as "fP," from entire PubMed abstracts using their MeSH IDs to extract articles more related to human diseases (see Table A2). The other articles were referred to as *other PubMed abstracts* (oP).

**Our BERT (fP + oP)**, or **ouBioBERT**, is our enhanced biomedical BERT model pre-trained from scratch using entire PubMed abstracts in which pre-training on medical articles, especially those related to human diseases, is amplified using our method. We used focused PubMed abstracts as the small corpus and other PubMed abstracts as the large corpus, and pre-trained the model using our method. For comparison with our model, we also pre-trained another model on entire PubMed abstracts using the conventional method from scratch, and named it *conv. bioBERT*.

## 2.4 Task-specific fine-tuning BERT

Given an input token sequence, a pre-trained language model generates an array of vectors in the contextual representations. A task-specific prediction layer is then placed on top to produce the final output for the task-specific application. Given the task-specific training data, the task-specific model parameters can be trained, and the BERT model parameters are fine-tuned by gradient descent using backpropagation. Figure 4 shows the general architecture of fine-tuning BERT models for downstream tasks. The input instance is first subjected to task-specific pre-processing and to the addition of special instance markers ([CLS], [SEP], etc.). The transformed input is then tokenized using the vocabulary of the neural language model and input into the neural language model. The sequence of vectors in contextual representations taken from the language model is then processed by Featurizer module and input into Predict module to produce the final output of the given task.

Three evaluations were performed. First, we studied the performance of the Japanese medical BERT variants and certain baseline models other than neural language models on a medical document classification task to confirm that our method can be used in Japanese. Second, we measured the BLUE benchmark scores of Our BERT (sP + sW) and pre-trained

*S.Wada et al.***Table 3.** BLUE tasks (Peng, et al., 2019).

| Corpus | Type | Task | Metrics | Domain |
| --- | --- | --- | --- | --- |
| MedSTS [5] | Sentence pairs | Sentence similarity | Pearson | Clinical |
| BIOSSES [8] | Sentence pairs | Sentence similarity | Pearson | Biomedical |
| BC5CDR-disease [10] | Mentions | Named-entity recognition | F1 | Biomedical |
| BC5CDR-chemical [10] | Mentions | Named-entity recognition | F1 | Biomedical |
| ShARe/CLEFE [16] | Mentions | Named-entity recognition | F1 | Clinical |
| DDI [17] | Relations | Relation extraction | micro F1 | Biomedical |
| ChemProt [3] | Relations | Relation extraction | micro F1 | Biomedical |
| i2b2 2010 [19] | Relations | Relation extraction | micro F1 | Clinical |
| HoC [20] | Documents | Document classification | F1 | Biomedical |
| MedNLI [21] | Pairs | Inference | Accuracy | Clinical |

BERT models using the conventional method with a single random seed to demonstrate the effectiveness of our method in English. Finally, we executed the BLUE benchmark with five different random seeds and compared the average score of Our BERT (fP + oP) with those of biomedical BlueBERT, clinical BlueBERT, and conv. bioBERT, respectively, to demonstrate the potential of our method.

## 3 Downstream tasks

### 3.1 Multiclass document classification task in Japanese

Because there is no shared task for medical-domain documents in Japanese, we created a multiclass document classification task using the medical topics found in the MSD Manual for the Professional [31] and named it DocClsJp. It comprises 2,475 articles that belong to one of the 22 disease categories.

We used the first 128 tokens of each document as an input sentence and defined its disease category as the correct label. We employed five-fold stratified cross-validation to evaluate the results based on micro-accuracy. To compare the BERT models, we also evaluated the performance of the conventional methods for DocClsJp.

### 3.2 BLUE benchmark

The BLUE benchmark, which comprises five different biomedical text-mining tasks with 10 corpora, was developed to facilitate research on language representations in the biomedical domain [14]. These 10 corpora are pre-existing datasets that have been widely used by the biomedical natural language processing community as shared tasks (see Table 3). Following the practice of Peng et al. (2019), we used a macro-average of F1-scores and Pearson scores to make comparisons among pre-trained BERT models as a total score. Moreover, to evaluate in detail the change in the total score by our method, we calculated the scores of the clinical and biomedical domains individually as clinical score and biomedical score, respectively. That is, the clinical score is the macro-average of MedSTS, ShARe/CLEFE, i2b2 2020, and MedNLI, and the biomedical score is the macro-average of BIOSSES, BC5CDR-disease/chemical, DDI, ChemProt, and Hallmarks of Cancer corpus (HoC) [3,5,8,10,16,17,19-21].

In this section, we briefly describe each of the individual tasks and datasets in the BLUE benchmark. For more information, refer to [14].

#### 3.2.1 Sentence similarity

The sentence similarity task is used to predict similarity scores based on sentence pairs. This can be handled as a regression problem. Therefore, a special [SEP] token is inserted between the two sentences, and a special [CLS] token is appended to the beginning of the input. The BERT encoding of [CLS] is used in the calculation of the regression score. We evaluated similarity using Pearson's correlation coefficients.

**BIOSSES** is a small dataset consisting of 100 pairs of sentences selected from the Text Analysis Conference Biomedical Summarization Track Training Dataset, which contains articles from the biomedical domain [8].

**MedSTS** is a dataset consisting of sentence pairs extracted from Mayo Clinic's clinical corpus and was used in the BioCreative/OHNLP Challenge 2018 Task 2 as ClinicalSTS [5].

#### 3.2.2 Named-entity recognition

The named-entity recognition task aims to recognize the mention spans given in a text. This is typically considered a sequential labeling task. The BERT encoding of a sequence of a given token is used to predict the label of each token and recognize mentions of entities of interest. We evaluated the predictions using the strict version of the F1-score. For disjoint mentions, all spans must also be strictly correct.

**BC5CDR-disease/chemical** is a dataset derived from the BioCreative V Chemical-Disease Relation corpus, which was produced to evaluate relation-extraction of drug-disease associated interactions [10]. We trained named-entity recognition models for disease (BC5CDR-disease) and disease (BC5CDR-chemical) individually.

The **ShARe/CLEF** eHealth Task 1 Corpus is a collection of clinical notes from the MIMIC II database [16]. Annotations are assigned to the disorders written in the clinical notes.

#### 3.2.3 Relation-extraction

The relation-extraction task predicts the relations and their types between the two entities mentioned in the sentences. Following the practice in the BLUE benchmark [14], we regard this task as a sentence classification task by anonymizing target named entities in a sentence using pre-defined tags such as @GENE$ and @CHEMICAL$ [13]. By replacing entity mentions with dummy tokens, overfitting can be avoided by memorizing the entity pairs.

**The DDI corpus** was developed for the DDI Extraction 2013 challenge and consists of 792 texts selected from the DrugBank database and 233 other MEDLINE abstracts [17].

**ChemProt** consists of PubMed abstracts with chemical-protein interactions between chemical and protein entities and was used for the BioCreative VI Chemical-Protein Interaction Track [3].

The **i2b2 2010** shared task was developed for the 2010 i2b2/VA Challenge to determine concepts, assertions, and relations in clinical texts. Annotations were given for the relationship between the medical problem and either the treatment, examination, or another medical problem.

#### 3.2.4 Document multilabel classification

The multilabel-classification task predicts multiple labels from texts. HoC was annotated with 10 hallmarks of cancer to help develop an automatic semantic classifier of scientific literature [20]. The text from PubMed abstracts was annotated at the sentence level. We followed the general practice and evaluated the example-based F1-score at the document level [14,32,33].

#### 3.2.5 Inference task

The inference task aims to predict whether the relationship between the premise and hypothesis sentences is a contradiction, entailment, or neutral.

**MedNLI** is an expert annotated dataset for NLI in the clinical domain and consists of sentence pairs sampled from MIMIC-III [26]. We measured the overall accuracy to evaluate the performance.

## 4 Experimental setups

For both pre-training BERT and fine-tuning it for downstream tasks, we leveraged mixed-precision training, called FP16 computation, which significantly accelerates the computation speed by performing operations in the half-precision format. We used two NVIDIA Quadro RTX 8000 (48 GB) GPUs for pre-training, whereas a single GPU was used for fine-tuning.

### 4.1 Pre-training BERT

We modified the implementation released by NVIDIA for training our models [34], which enabled us to leverage FP16 computation, gradient accumulation, and a layer-wise adaptive moments based (LAMB) optimizer [35]. The pre-training configuration was almost the same as that of BERT-Base unless stated otherwise.

#### 4.1.1 Japanese medical BERT from a small medical corpus

For Our BERT (jpCR + jpW) and BERT (jpCR), the maximum sequence length was fixed at 128 tokens, and the global batch size (GBS) was set to 2,048. Additionally, LAMB optimizer with a learning rate (LR) of 7e–4 was used. We trained the model for 125K steps. The size of the vocabulary was 32K. BERT (jpW/jpCR) was initialized from BERT (jpW) and trained using jpCR until the loss of MLM and NSP on the training dataset stopped decreasing. Additionally, we used LAMB optimizer with an LR of 1e–4.

#### 4.1.2 English biomedical corpus from a small biomedical corpus

We used the same set of settings for Our BERT (sP + sW), BERT (sW), and BERT (sP) as that for Our BERT (jpCR + jpW) or BERT (jpCR). BERT (sW/sP) was initialized from BERT (sW) and trained using sP until the loss of MLM and NSP stopped decreasing with the same settings of the maximum sequence length and GBS as that for BERT (jpW/jpCR).

#### 4.1.3 Enhanced biomedical BERT from whole PubMed abstracts

**Table 4.** Range of the number of training epochs for each task/dataset.

| Dataset | Number of epochs |
| --- | --- |
| MedSTS | {7, 8, 9, 10} |
| BIOSSES | {40, 50} |
| Named-entity recognition | {20, 30} |
| Relation-extraction | {5, 6, 7, 8, 9, 10} |
| HoC | {5, 10, 15} |
| MedNLI | {5, 6, 7, 8, 9, 10, 15} |
| DocClsJp | {3, 4, 5, 6, 7, 8, 9, 10} |

For Our BERT (fP + oP), we followed NVIDIA's implementation. First, we set the maximum sequence length of 128 tokens and trained the model for 7,038 steps using GBS of 65,536 and LAMB optimizer with an LR of 6e–3. Subsequently, we continued to train the model, allowing a sequence length of up to 512 tokens for an additional 1,563 steps to learn positional embeddings using GBS of 32,768 and LAMB optimizer with an LR of 4e–3. The size of the amplified vocabulary was 32K. For conv. bioBERT, we used the same settings as those used in Our BERT (fP + oP), except using the conventional method.

### 4.2 Fine-tuning BERT for downstream tasks

We mostly followed the same architecture and optimization provided in transformers for fine-tuning [29]. In all settings, we set the maximum sequence length to 128 tokens and employed Adam [36] for fine-tuning using a batch size of 32 and an LR of 3e–5, 4e–5, or 5e–5, respectively. The number of training epochs was set for each task, as shown in Table 4. For each dataset and BERT variant, we selected the best LR and number of epochs on the development set, and then reported the corresponding test results.

### 4.3 The performance of the baseline in DocClsJp

To evaluate the performance of the baseline, several conventional methods were applied.

A classical method for text classification tasks is to use support vector machines (SVM) to classify documents with features obtained from them [6]. The features are based on TF-IDF, numerical statistics that indicate the importance of a word in a text by scoring the words in the document, considering the corpus to which the document belongs.

Deep neural networks for text classification tasks used before the introduction of transformer-based language models include convolutional neural networks (CNNs) and bidirectional long short-term memory (biLSTM) with self-attention [7,9]. We first learned the word embeddings of the Japanese clinical references using fastText [37]. Consequently, we converted a sequence of words from the documents using the embeddings and fed it into their neural networks. The structures of their networks were prepared based on the architecture of their original papers [7,9].

For the three baseline methods, the maximum length of the input was set to 128 to match the input of our BERT models. The optimal hyperparameters were found using Optuna, a hyperparameter optimization software designed using the define-by-run principle [38].

## 5 Results

**Table 5.** Test results on DocClsJp.

| Model | F1-score |
|---|---|
| TF-IDF + SVM | 68.8 (1.3) |
| CNN | 77.3 (2.8) |
| biLSTM with SA | 78.9 (1.8) |
| BERT (jpW) | 82.3 (1.9) |
| UTH-BERT (EMR) | 82.7 (1.1) |
| BERT (jpCR) | 84.4 (2.6) |
| BERT (jpW/ jpCR) | 84.6 (2.6) |
| Our BERT (jpCR + jpW) | |
|   SimPT    AmpV | |
|   ✓          ✓ | **87.2 (1.3)** |
|   ✓ | 85.6 (2.4) |

*Notes:* The numbers represent the means (standard deviations) obtained using five-fold stratified cross-validation. TF-IDF + SVM: Support Vector Machines with TF-IDF [6]; CNNs: Convolutional Neural Networks for sentence classification [7]; biLSTM with SA: bidirectional Long Short-Term Memory with self-attention [9]; SimPT: simultaneous pre-training after up-sampling; AmpV: amplified vocabulary.

Table 5 compares the micro-accuracy of the model pre-trained using our method and those of the others on DocClsJp. The performance of the BERT variants was higher than that of the baseline models of the other three. Our results showed a higher performance of Our BERT (jpCR + jpW) than those of the other pre-trained models (either constructed using known techniques or publicly released). BERT (jpCR) had a higher score than BERT (jpW) and the same score as BERT (jpW/jpCR). The ablation tests showed that SimPT was more effective than existing methods and that its performance was enhanced by modifying the vocabulary with our method.

Table 6 summarizes the performance of Our BERT (sP + sW) in terms of the BLUE score. In contrast to the experiment in Japanese, each score tended to be lower for BERT (sP) than for BERT (sW). For Our BERT (sP + sW), a major number of individual scores in each dataset were higher than those obtained using conventional methods. Consequently, our model outperformed the other three models. The ablation tests showed that AmpV was as effective as our Japanese model.

Table 7 compares the summarized score of Our BERT (fP + oP) on the BLUE benchmark with those of biomedical BlueBERT, clinical BlueBERT, and conv. bioBERT. Conv. bioBERT showed a higher score than the published models. Clinical BlueBERT, which was initialized with biomedical BlueBERT and additionally pre-trained on MIMIC-III clinical notes, had the highest clinical score; however, the biomedical score was considerably lower. On the other hand, Our BERT (fP + oP) did not show

**Table 7.** Performance of Our BERT (fP + oP) and its ablation tests on the BLUE task.

| Model | Total score | Clinical score | Biomedical score |
|---|---|---|---|
| biomedical BlueBERT | 82.9 (0.1) | 79.8 (0.2) | 85.0 (0.1) |
| clinical BlueBERT | 81.6 (0.5) | **81.0** (0.3) | 81.9 (0.9) |
| conv. bioBERT | 83.6 (0.1) | 80.2 (0.3) | 85.8 (0.2) |
| Our BERT (fP + oP) | | | |
|   SimPT    AmpV | | | |
|   ✓          ✓ | **83.9** (0.2) | 80.5 (0.2) | **86.1** (0.2) |
|   ✓ | 83.9 (0.3) | 80.6 (0.1) | 86.0 (0.4) |

*Notes:* The numbers represent the means (standard deviations) on five different random seeds. The best scores are written in bold, and the second best are underlined. SimPT: simultaneous pre-training after up-sampling; AmpV: Amplified vocabulary.

a decrease in the biomedical score while the clinical score increased; consequently, its total score was the highest of the four models. We also conducted ablation tests and found that SimPT was especially successful in improving clinical scores as well as biomedical scores. In this experiment, AmpV was not effective. Detailed results are presented in Table A3.

## 6 Discussion

We confirmed that the models trained using our method proved robust on the BLUE benchmark even when using a small medical corpus, and we demonstrated that our method could construct both localized medical BERT and enhanced biomedical BERT. These results suggest the importance of adapting the corpus used for pre-training to the target task, and the effectiveness of our proposed method in supporting it.

We first applied our method to the medical BERT in Japanese and evaluated it for a single task. In the experiment, Our BERT (jpCR + jpW) outperformed both the baseline models and the other BERT variants. Furthermore, in the ablation study, we observed that the performance improved when using the customized vocabulary via our method. Interestingly, UTH-BERT (EMR), which was designed for the clinical domain in Japanese, was as accurate as BERT (jpW), which was for the general domain. This is likely because DocClsJp is a classification task for medical references and the corpus used for pre-training Our BERT (jpCR + jpW)

**Table 6.** BLUE scores of Our BERT variants.

| Model | Total | Med STS | BIOSSES | BC5CDR -disease | BC5CDR -chemical | ShARe/ CLEFE | DDI | Chem Prot | i2b2 2010 | HoC | MedNLI |
|---|---|---|---|---|---|---|---|---|---|---|---|
| BERT (sW) | 79.3 | 82.7 | 90.0 | 82.3 | 90.6 | 75.4 | 76.7 | 64.8 | 68.9 | 84.6 | 76.6 |
| BERT (sP) | 77.9 | 78.8 | 80.5 | 84.7 | 90.8 | 76.4 | 74.5 | 63.6 | 68.2 | 84.2 | 77.4 |
| BERT (sW/sP) | 79.7 | 83.7 | 89.7 | 83.6 | 91.4 | 77.3 | 73.8 | 65.4 | 69.9 | 84.8 | 77.9 |
| Our BERT (sP + sW) | | | | | | | | | | | |
|  SimPT   AmpV | | | | | | | | | | | |
|  ✓       ✓ | **81.1** | 83.0 | 91.1 | **84.9** | 91.4 | 77.4 | **78.0** | 67.3 | 73.2 | **85.2** | 79.6 |
|  ✓ | 80.3 | **84.2** | **91.7** | 84.3 | 91.2 | 77.3 | 75.1 | 66.2 | 70.4 | 84.6 | 78.1 |

*Notes:* The best scores are written in bold, and the second best are underlined.

BERT (sW) is a model for use in the general domain. BERT (sP) and BERT (sW/sP) are the models that are pre-trained via the conventional method using a small biomedical corpus.

consists of clinical references and is therefore similar to the domain of the task. Similar results have been observed in English when comparing between BioBERT constructed from PubMed and clinicalBERT using MIMIC-III clinical notes [13,15].

Next, to simulate the experiment in Japanese, we created Our BERT (sP + sW) by combining a small biomedical corpus and a large general corpus in English. It performed sufficiently well for practical use. BERT (sP), which was pre-trained using only small PubMed abstracts, and BERT (sW/sP), which was initialized from BERT (sW) and pre-trained using only small PubMed abstracts, performed worse than Our BERT (sP + sW). These results suggest that the pre-trained models using conventional methods suffer from overfitting and that our method can avoid this issue. This supports the effectiveness of our method in using a small corpus in English. Therefore, it might be applicable to other languages as well. Furthermore, our method can also be applied to professional domains other than the medical domain.

Finally, we found that a high-performance pre-trained model can be trained using our method with Our BERT (fP + oP). The result of conv. bioBERT identified that the configuration we used in the pre-training of BERT models was the most significant factor responsible for the improvement in their scores. Previous studies have reported that larger batch sizes and longer steps for pre-training are effective in improving performance [35,39]; therefore, our model is likely to benefit from them. Furthermore, our SimPT achieved an improvement in the BLUE benchmark scores, especially in clinical scores, although we used only PubMed abstracts rather than clinical notes. The results for clinical BlueBERT showed that the corpus used for the last pre-training had the greatest impact on performance and that its high clinical score came at the expense of the biomedical score. The success of our model shows that enhanced pre-training of biomedical articles close to the specific-task can improve model performance even when the available language resources are limited. On the other hand, there was a difference in the ablation study between the third experiment and the previous two; the usefulness of AmpV could not be confirmed. This might be due to the difference in vocabularies between "with" and "without" AmpV. The differences were 25.8 %, 29.8 %, and 10.3 % in Our BERT (jpCR + jpW), BERT (sP + sW), and BERT (fP + oP), respectively. We assumed that there was no obvious difference in the third result because the difference was smaller for the latter than for the former two.

The common denominator of the three experiments was the scarcity of resources used for pre-training suitable for their target tasks. It is clear that pre-training on corpora corresponding to target tasks would be needed; however, no solution to the lack of such corpora has been shown. While the release of PubMedBERT raised some questions about the need to address both clinical and biomedical tasks [40], we demonstrated that our up-sampling method could serve as an option to solve this problem by constructing ouBioBERT.

This study had several notable limitations. First, we checked the robustness of our models on multiple tasks in English; however, we evaluated Our BERT (jpCR + jpW) for a single task in Japanese. This is because there are no text-mining shared tasks in Japanese for the medical domain, and it is difficult to directly solve this problem. Second, our proposed method might be applicable to other languages as well; however, we did not conduct additional studies to confirm this. Such studies are highly computationally expensive and significantly time-consuming in our environment, although their experiments would reinforce our method.

## 7 Conclusions

We introduced a pre-training technique that consists of simultaneous pre-training after up-sampling with an amplified vocabulary and confirmed that we could produce high-performance BERT models that deal with targeted tasks. We first showed that a practical medical BERT model can be constructed via our method using a small medical corpus in Japanese and that it could then be applied in English. Additionally, using ouBioBERT, we confirmed that a pre-trained biomedical model that also managed clinical tasks can be produced using our method. These results support the validity of our hypotheses. Our study could help overcome the challenges of biomedical text-mining tasks in both English and other languages.

## Abbreviations

AmpV: amplified vocabulary; BC5CDR: BioCreative V Chemical-Disease Relation corpus; BERT: bidirectional encoder representations from transformers; biLSTM: bidirectional long short-term memory; BLUE: biomedical language understanding evaluation; BPE: byte-pair encoding; CNNs: convolutional neural networks; EMR: electronic medical record of the University of Tokyo Hospital; fP: focused PubMed abstracts; GBS: global batch size; jpCR: Japanese clinical refer-ences; jpW: Japanese Wikipedia; LAMB: layer-wise adaptive moments based; LR: learning rate; M: MIMIC-III clinical notes; MeSH: medical subject headings; MLM: masked language modeling; NLI: natural language inference; NLP: natural language processing; NSP: next-sentence pre-diction; oP: other PubMed abstracts; P: PubMed abstracts; PDF: portable document format; simPT: simultaneous pre-training after up-sampling; sP: small PubMed abstracts; SVM: support vector machines; sW: sampled English Wikipedia

## Acknowledgments

Not applicable.

## Authors' contributions

SW designed the project, developed the models and the codes, and was a major contributor in writing the manuscript. YM acquired the financial support, and supervised the project. JK investigated and analyzed the baseline. TT, SM, SK, and YM provided substantial contributions during manuscript writing and revision. All authors read and approved the final manuscript.

## Funding

This work was supported by the Council for Science, Technology and Innovation, Cross-ministerial Strategic Innovation Promotion Program, "Innovative AI Hospital System" (Funding Agency: National Institute of Biomedical Innovation, Health, and Nutrition). The funding body had no role in the design of the study; collection, analysis, and interpretation of data; or in writing the manuscript.

## Availability of data and material

Our BERT (fP + oP) (named *ouBioBERT*), along with the source code for fine-tuning, is available freely at https://github.com/sy-wada/blue_benchmark_with_transformers. The pre-trained weights of Japanese medical BERT models in this study are available for academic purposes at https://github.com/ou-medinfo/medbertjp. The DocClsJp dataset is not publicly available because of restrictions on secondary distribution of copyrighted works, but it is available from the corresponding author upon reasonable request.

## Ethics approval and consent to participate

Not applicable.

## Consent for publication

Not applicable.

## Competing interests

The authors declare that they have no competing interests.

# 8 Appendix

Table A1. List of the names for our models and the published models discussed in this study.

| Model name | Characteristics of pre-training |
|---|---|
| *English* | |
| BlueBERT-BASE (P) <br> > biomedical BlueBERT | 1. English Wikipedia + BooksCorpus <br> 2. PubMed abstracts <br>    via the conventional method |
| BlueBERT-BASE (P + M) <br> > clinical BlueBERT | 1. English Wikipedia + BooksCorpus <br> 2. PubMed abstracts <br> 3. MIMIC-III clinical notes <br>    via the conventional method |
| BERT (sW) | 1. sampled English Wikipedia <br>    via the conventional method |
| BERT (sP) | 1. small PubMed abstracts <br>    via the conventional method |
| BERT (sW/sP) | 1. sampled English Wikipedia <br> 2. small PubMed abstracts <br>    via the conventional method |
| Our BERT (sP + sW) | 1. small PubMed abstracts <br>    + sampled English Wikipedia <br> via **our method** |
| conv. bioBERT | 1. entire PubMed abstracts <br> via the conventional method |
| Our BERT (fP + oP) | 1. focused PubMed abstracts <br>    + other PubMed abstracts <br> via **our method** |
| *Japanese* | |
| TOHOUK-BERT <br> > BERT (jpW) | 1. Japanese Wikipedia <br>    via the conventional method |
| UTH-BERT <br> > UTH-BERT (EMR) | 1. electronic medical records <br>    via the conventional method |
| BERT (jpCR) | 1. clinical references <br>    via the conventional method |
| BERT (jpW/jpCR) | 1. Japanese Wikipedia <br> 2. clinical references <br>    via the conventional method |
| Our BERT (jpCR + jpW) | 1. clinical references <br>    + Japanese Wikipedia <br> via **our method** |

*Notes:* Parentheses indicates the corpora used in its pre-training. A "/" symbol indicates an additional pre-training. A "+" symbol indicates that the simultaneous pre-training was performed after up-sampling.

For example: "BERT (jpW/jpCR)" indicates that it was pre-trained first using the Japanese Wikipedia and also using clinical references.

Abbreviations. P: PubMed abstracts; M: MIMIC-III clinical notes; sW: sampled English Wikipedia; sP: small PubMed abstracts; fP: focused PubMed abstracts; oP: other PubMed abstracts; jpW: Japanese Wikipedia; EMR: electronic medical record of the University of Tokyo Hospital; jpCR: clinical references.

Table A2. Selection of target articles using the MeSH tree number for sP and fP.

| Flag | MeSH Tree Number | small PubMed abstracts (sP) | Focused PubMed abstracts (fP) |
|---|---|---|---|
| **Included** | [C] Diseases | ✓ | ✓ |
| **Excluded** | [A13] Animal Structures | ✓ | |
| | [A16] Embryonic Structures | ✓ | |
| | [A18] Plant Structures | ✓ | ✓ |
| | [A19] Fungal Structures | ✓ | ✓ |
| | [A20] Bacterial Structures | ✓ | ✓ |
| | [A21] Viral Structures | ✓ | ✓ |
| | [B01.650] Plants | ✓ | ✓ |
| | [B02] Archaea | ✓ | ✓ |
| | [B03] Bacteria | ✓ | |
| | [B04] Viruses | ✓ | |
| | [B05] Organism Forms | ✓ | ✓ |
| | [C22] Animal Diseases | ✓ | |
| | [D26] Pharmaceutical Preparations | ✓ | |
| | [E03] Anesthesia and Analgesia | ✓ | |
| | [E05] Investigative Techniques | ✓ | ✓ |
| | [E06] Dentistry | ✓ | |
| | [E07] Equipment and Supplies | ✓ | ✓ |
| | [F02] Psychological Phenomena | ✓ | ✓ |
| | [F04] Behavioral Disciplines and Activities | ✓ | ✓ |
| | [G] Phenomena and Processes | ✓ | |
| | [G17] Mathematical Concepts | | ✓ |
| | [H01] Natural Science Disciplines | ✓ | ✓ |
| | [I] Anthropology, Education, Sociology, and Social Phenomena | ✓ | ✓ |
| | [J] Technology, Industry, and Agriculture | ✓ | ✓ |
| | [K] Humanities | ✓ | ✓ |
| | [L] Information Science | ✓ | ✓ |
| | [N] Named Groups | ✓ | ✓ |
| | [Z] Geographicals | ✓ | ✓ |

If an article has an "Included" tree number but does not have an "Excluded" one, it is considered a target article. Furthermore, to adjust the corpus size of sP, we sampled articles published after 2010.

Table A3.  Performance of ouBioBERT on the BLUE task in detail.

| Model | | Total | MedSTS | BIOSSES | BC5CDR -disease | BC5CDR -chemical | ShARe/ CLEFE | DDI | ChemProt | i2b2 2010 | HoC | MedNLI |
|---|---|---|---|---|---|---|---|---|---|---|---|---|
| biomedical BlueBERT | | 82.9 (0.1) | 84.8 (0.5) | 90.3 (2.0) | 86.2 (0.4) | 93.3 (0.3) | 78.3 (0.4) | 80.7 (0.6) | 73.5 (0.5) | 73.9 (0.8) | **86.3** (0.7) | 82.1 (0.8) |
| Clinical BlueBERT | | 81.6 (0.5) | 84.6 (0.8) | 82.0 (5.1) | 84.7 (0.3) | 92.3 (0.1) | **79.9** (0.4) | 78.8 (0.8) | 68.6 (0.5) | **75.8** (0.3) | 85.0 (0.4) | **83.9** (0.8) |
| conv. bioBERT | | 83.6 (0.1) | **85.0** (0.3) | 92.6 (0.6) | 87.1 (0.3) | **94.1** (0.3) | 78.9 (0.6) | 81.1 (0.7) | 74.1 (0.3) | 74.1 (0.6) | 85.9 (0.4) | 82.7 (0.5) |
| ouBioBERT | | | | | | | | | | | | |
| *SimPT* | *AmpV* | | | | | | | | | | | |
| ✓ | ✓ | **83.9** (0.2) | 84.5 (0.5) | 93.1 (0.5) | 87.3 (0.3) | 93.9 (0.1) | 79.8 (0.3) | 80.7 (0.7) | **75.5** (0.3) | 74.7 (0.5) | 86.2 (0.6) | 83.2 (0.4) |
| ✓ | | 83.9 (0.3) | 85.0 (0.2) | **93.3** (1.4) | **87.4** (0.2) | 93.9 (0.4) | 79.4 (0.4) | **81.3** (0.9) | 74.8 (0.3) | 74.6 (0.5) | 85.4 (0.8) | 83.6 (0.4) |

*Notes:* The numbers represent the means (standard deviations) on five different random seeds. The best scores are written in bold, and the second best are underlined. SimPT: simultaneous pre-training after up-sampling; AmpV: Amplified vocabulary.